# Feature Concepts for Data Federative Innovations

Yukio Ohsawa[1], Sae Kondo[2], Teruaki Hayashi[1]

1: Dept. Systems Innovation, School of Engineering
The University of Tokyo
`ohsawa@sys.t.u-tokyo.ac.jp`

2: Dept. Architecture, Graduate School of Engineering,
Mie University
`skondo@arch.mie-u.ac.jp`

**Abstract.** A *feature concept*, the essence of the data-federative innovation process, is presented as a model of the concept to be acquired from data. A feature concept may be a simple feature, such as a single variable, but is more likely to be a conceptual illustration of the abstract information to be obtained from the data. For example, trees and clusters are feature concepts for decision tree learning and clustering, respectively. Useful feature concepts for satisfying the requirements of users of data have been elicited so far via creative communication among stakeholders in the market of data. In this short paper, such a creative communication is reviewed, showing a couple of applications, for example, change explanation in markets and earthquakes, and highlight the feature concepts elicited in these cases.

## 1       Introduction:

Logically, the process of thoughts and communication in Innovators Marketplace on Data Jackets (IMDJ) we executed so far can be modeled on the framework of abductive reasoning (Ohsawa et al., 2017). Using this model, the necessity to elicit information about contexts, that is, situations where to use data and/or receive the services or products created based on data, has been positioned as a key scope in creating a solution for satisfying a requirement. Here, a context can be regarded as the situation of a data user or a data subject, to be obtained by asking why a requirement has been proposed and how a solution can be realized. For example, if the requirement is to reduce traffic congestion, the reason (answer to "why") may be the context in which people desire to avoid pandemics. If so, one can suggest that the solution for realizing this requirement may be to use the data on human flow, population, effective reproduction rate, etc. On the other hand, if the solution is the sheer reduction of traffic congestion, the method (answer to "how") may be to combine the datasets on road conditions and congestion. Thus, contextual information plays an essential role in embodying a data-based solution. In this sense, we can say asking deep reasoning questions (why) and generative design questions (how) contribute significantly to data-federative innovations, that is, innovations by strategic combination of datasets, as well as to designing [Eris 2004].

The limit of the IMDJ often comes from the gap between the expressed requirements reflecting humans' subjective desires or thoughts and the data to be used, which do not reflect the subjective information. Although DJs accept subjective descriptions about the expectations and desires of data providers or individuals with knowledge about data, the DJs are insufficient because of the bias of the subjectivity of each individual in comparison with the various interests of the entire society. Therefore, so far, we urged participants in the IMDJ to speak out subjective opinions so that we could revise the DJs reflecting the opinions for future sessions. However, even though we urge participants to open their minds to others, they still suffer from a missing link between data and the requirement, that is, the abstract image about the expected information to be acquired using datasets for the requirement satisfaction.

## 2 Feature concepts for connecting requirements and data

A *feature concept* is an abstract image of the information or knowledge to be acquired by using data linked to the method i.e., how, the reason i.e., why, and the dataset(s) i.e., what, should be used to satisfy a requirement. In the examples shown below, we discover that creativity in IMDJ has been enhanced by and for eliciting, using, and sharing feature concepts in various forms. Each form could be a new variable, a new predicate, a new function, or even a new logical clause (if the participants could express concepts in the logical framework), but can be more generally illustrated as images in Fig.1. For example, unsupervised machine learning methods such as clustering with cutting noise events [Fränti and Yang 2018, Cheng, D.,et al. 2019] or SOINN [Shen and Hasegawa 2006] are algorithms where a hidden cluster is restored from data including scattered noise signals. Thus, *embedded clusters*, perceived as the desired information to be acquired from the data, can be interpreted as the feature concept, as shown in Fig.1(a), if the user's requirement is to extract hidden structures behind noisy data, for example, data on epicenter positions. As well, decision tree learning is used to extract *trees,* as shown in Fig.1(b), to classify entities in the data to explain a given class, and the random forest [Ho 1995] means extracting a *forest*.

In these methods, trees and forests are used as feature concepts, suiting the requirements for classifying samples and obtaining multiple trees for a more precise classification than a single tree. If a forest is not intuitive to a user, such a feature concept as a *conference* may be better to use where the idea of each participant corresponds to a tree. On the other hand, association rules (Agrawal and Srikant 1994) can be represented by a feature concept *with a set of nodes connected via arrows*, where the nodes at the tail represent the conditional items or events and the nodes at the head indicate the conclusional ones that tend to occur if the conditional ones do.

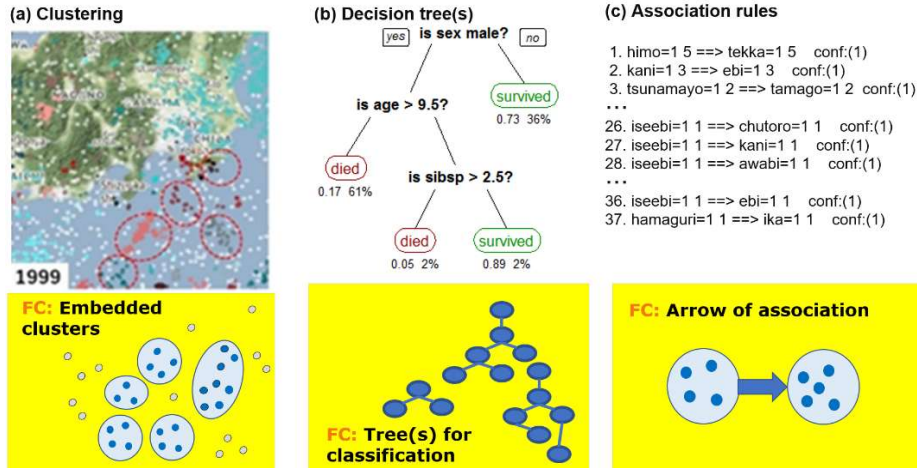

**Fig. 1.** Examples of feature concepts for three basic methods for data mining.

Some tools share the same feature concept, although the algorithms may vary owing to the various efforts of the researchers who refined the previous work. In these efforts, other feature concepts were exploited. For example, in the case of decision tree, the variables to be used at a higher level such as the gender ("sex" in the figure) of an individual to explain the survival rate in the accident of Titanic is selected on the variable's quantity of information i.e., maximum reduction of the entropy, since the earliest version ID3 [Quinlan 1986] to more recent ones such as random forest [Ho 1995]. This use of entropy can be interpreted as finding the vessel in which the victimized individuals are distributed more frequently than out of which. This *vessel* may be regarded as a feature concept that fits the requirement to select a useful variable.

Feature concepts, if provided explicitly, play an important role in bridging social requirements and features in datasets because they originate from the requirements of data users and can be projected onto the methods, tools, and algorithms to be used for taking advantage of the data utility, as shown [Ohsawa et al. 2022]. As shown in Fig.2, a feature concept may be given by some other than the creator of an algorithm if the purpose is to foster communication about the tool in the data market. Thus, contexts and feature concepts connect elements of the data market – datasets, knowledge, and skills for using/reusing data – for data-federative innovation.

In addition to a frame of desired knowledge or of the thoughts or communication about the knowledge, the feature concept can be used as the performance dimension of data analysis because it is a feature desired to be extracted from data. Thus, a feature concept is useful for both creating and evaluating solutions, as well as for realizing scientific communication to satisfy requirements that consider data.

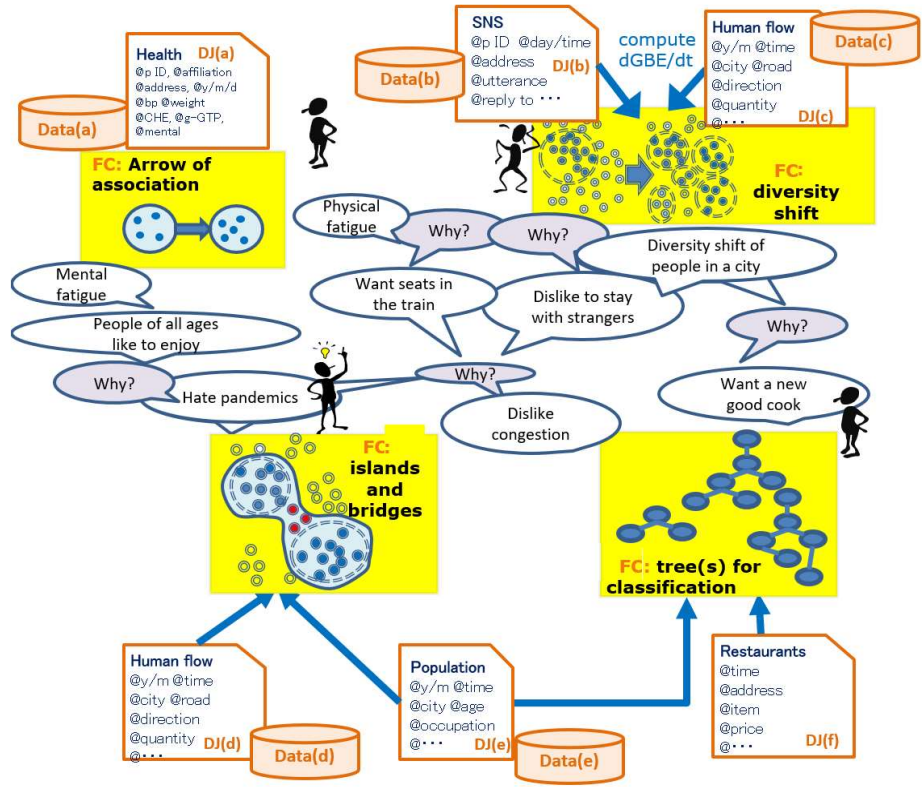

**Fig. 2.** The images and positions of feature concepts (FC through FC99 here) in the communication to connect requirements to solutions and DJs.

## 3  Examples of feature concepts in data utilities

The next example of the feature concept is the development of KeyGraph [Ohsawa et al. 1998] by the author. KeyGraph was first developed as a method for indexing, that is, extracting keywords from a document (D1-1 below: DJs were not invented yet, so let us put just D instead of DJ), and is nowadays used as a tool for information visualization systems to aid in the process of creative decision making [Ohsawa 2003, Hong et al. 2006, Fruchter et al. 2005] from sequential data (D1-2).

At the time of its first development, the context was the interest of natural language researchers in extracting a rare keyword, that is, a low-frequency word that carries the essence of the contextual flow in the target document. For example, in Fig.3, the visualization of word-word correlations is shown using KeyGraph. The story of The Arrest of Lupin changes the context from a peaceful journey on a boat to the mood of anxiety due to a wireless telegraph reporting the information that Arsene Lupin is on board. However, "wireless" appears only three times and "telegraph" just twice, whereas "Lupin" the rubber and "Ganimard" the detective appeared 37 and 11 times respectively in spite of the small information they carry -- Lupin and Ganimard appears quite usually in the novel series of Arsense Lupin by Leblanc. The problem was how to extract "wireless," "telegraph," etc., as keywords in this text. The tfidf criteria evaluated the importance of "wireless telepgraph" even less than the sheer frequency. The success of KeyGraph in Fig.3 was because this algorithm was created by responding to the requirement and the idea (solution below) for the datasets as follows.

**Example 1: KeyGraph on the architecture model** [Ohsawa et al 1998]
**Req1:** Low-frequency words or items representing the essence of contextual flow in sequential data
**Sol1**: Visualize "bases" and "roofs" in the metaphor regarding the target data (text) as an architecture of a building, and find important roofs
**D1-1**: text (document)
**D1-2**: A sequence (items in a supermarket, earthquakes: applied later than DJ1, not combined but independent of DJ1)

KeyGraph has been proposed based on a simple model called the architecture model, which comes from a metaphor in which a document is compared to a constructed building. This building has bases (statements for preparing basic concepts), columns, walls, and windows (functions to protect people inside). However, after all, the roofs (main ideas in the document), without which the building's inhabitants cannot be protected against rain or sunshine are the most important. The roofs were supported by columns on the bases. To extract the roofs, KeyGraph is composed of three major phases.

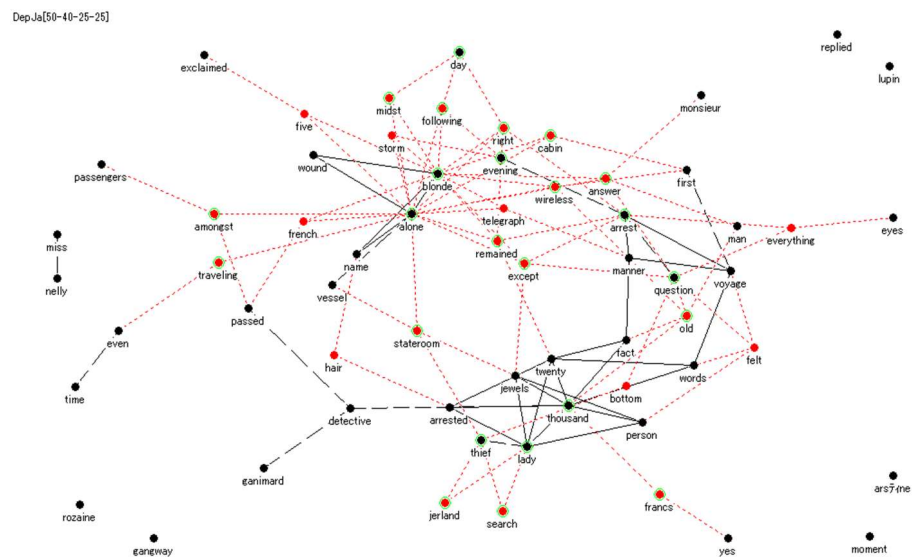

**Fig. 3.** The result of KeyGraph applied to The Arrest of Arsene Lupin (Leblanc, M. 1905)

1) Extracting the bases, where basic and preparatory concepts are obtained as clusters, each of which are formed as clusters on the cooccurrence among words
2) Extracting columns, where columns, i.e., the relationships between words in the document and the bases extracted above are obtained.
3) Extracting roofs, which words at the crosses of strong columns are regarded as.

Making a co-occurrence graph in 1) was already a common approach in 1998 for computing the relevance between terms. The new point of KeyGraph was to regard the co-occurrence graph as the basis of the document, on which the ideas in the document are built in a hierarchical manner. Thus, the result of KeyGraph consists of the following objects, replacing "word" with "item" in order to generalize for later applications.

- Black nodes indicate the items in the *bases* that frequently occur in a dataset.
- White (or red in a revised version) nodes indicating the *roofs*, not occurring as frequently as the black nodes but co-occurring with black nodes in the dataset.
- Double-circled nodes indicating items co-occurring frequently with the items of black nodes, regarded as keywords. These are the most important roofs.
- Edges indicating frequent co-occurrence in dataset Solid (black) edges are used to form *bases*, and dotted (red) lines are used to connect the *bases* via *roofs*.

The *architecture* can be regarded as a feature concept. The difference of a feature concept from just a model for analysis is that a feature concept comes from human ideas originating from thoughts about requirements, which may have to be transformed into a computable model if necessary. In the above case, the architecture model originated from a requirement to grasp important words in spite of their low frequency, and is simply put into a computation procedure. On the other hand, when we diverted KeyGraph to chance discovery, that is, to detect and explain an event having a potential importance to human decision-making, a new feature concept has been proposed, that is, *islands and bridges* [Ohsawa 2003]. For example, suppose the target is POS data of supermarket as in D1-2 above, the composite parts to be visualized by KeyGraph can be as follows. The two concept features are shown in Fig.4.

- An *island*, a cluster consisting of black nodes linked by solid lines, corresponding to the basis of the architecture model. For example, the set of snack items as {"chips," "cheese snack," "cracker"} forms one island, and liquor items {"beer," "wine," "sake"} forms another.

- A *bridge* is defined as dotted lines connecting islands. Containing nodes not belonging to any island, as relay points of dotted lines, are represented by red nodes such as {"caviar"} that appear at a lower frequency than the items on islands.

The islands can be viewed as the underlying common contexts of consumers' lives, because they are formed by a set of items co-occurring frequently in the data set. In the market example here, a customer visiting the supermarket to buy beer and wine may be interested in caviar, but gives up buying it considering the price and moves to the shelf of snacks to buy chips and crackers. In this case, the caviar worked as a bridge that urged the customer from the island of liquor to the island of something to eat.

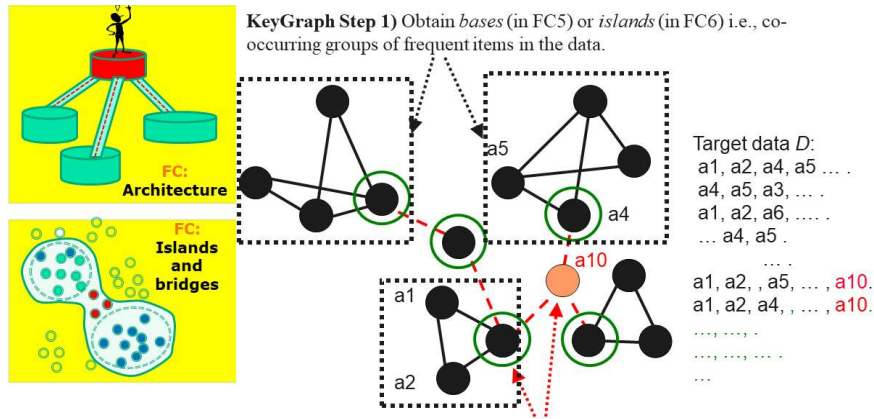

**KeyGraph Step 1)** Obtain *bases* (in FC5) or *islands* (in FC6) i.e., co-occurring groups of frequent items in the data.

**KeyGraph Step 2)** Obtain *roofs* (in FC5) or *bridges* (in FC6), i.e., items co-occurring with multiple bases or islands respectively. If the node is of lower frequency than black nodes (e.g. a10), it is a new node put as a red one. Otherwise it is a black node surrounded by circles (e.g. a4). And, these nodes represent the candidates of chances in chance discovery.

**Fig. 4.** Two feature concepts of KeyGraph

Here, we review a case presented in (Ohsawa and Usui 2005), where a map obtained by KeyGraph assisted chance discovery in business, as shown in Fig.5. The dataset was of a "pick-up" sequence in a textile exhibition organized by textile company A, where visitors representing apparel firms and textile-converter firms came to pick up interesting items among about 800 textile samples. Each order card was written manually, showing a set of items picked up by one visitor, and a dataset of as many lines as the number of cards (hundreds) was collected. KeyGraph was applied to this dataset.

Ten marketers in textile firm A attached real textile samples to the printed map visualized by KeyGraph, to look at the graph structure and touch the textile items on it and discuss their plans about whom, why, and how to sell the textiles. First, they noticed the islands corresponding to popular item sets: the large island in the right meaning an established market of textiles for business clothes (suits and shirts for under suits), and the one on the left made of a single node regarded as an island in the sense it is frequent. The latter small island was interpreted as a textile for casual wear by multiple marketers. One of the marketers pointed out that consumers desire to change from the island in the right to the left. That is, when a working person moves from the workplace to a restaurant for dinner after working, one may like to change clothes from business suits to casual for relaxation. Interested in this scenario of consumers' behavior, the marketers came to pay attention to the infrequent items between the islands, that is, to the dotted red ellipse. As a result, the marketers created a new scenario to sell the new corduroy in the red ellipse to suit the manufacturer as a material of casual jacket for office workers to wear in going out for dinner. In this example, the dotted red ellipse shows a chance because it affected the decision of the marketers and expected consumers of the new casual jacket. We can say that the marketers worked as conceptualized in the feature concept of *islands and bridges*.

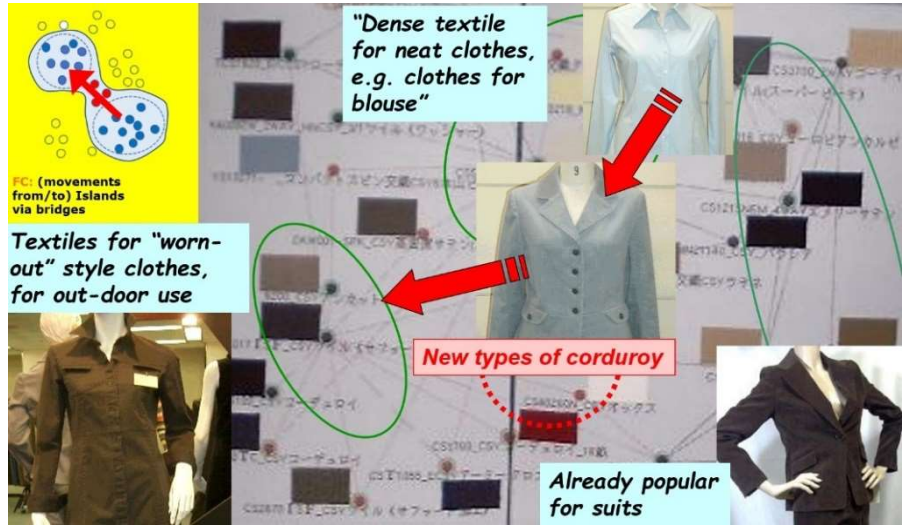

**Fig. 5.** An realization of the feature concept *islands and bridge* using KeyGraph (Ohsawa and Nishihara 2012)

On the other hand, Example 2 was realized when the author introduced a feature concept "*change explanation*" and "*diversity shift*." Sol2 was embodied later than the presentation of the original solution, *which explains changes in the market by showing causal events such as items or behaviors of customers*."

**Example 2: Change explanation in businesses**
**Req2:** Detect and explain causal events in the change points in the market
**Sol2**: Explain changes in the market with visualized "*explanatory changes*" implying the latent dynamics such as the "*trend shift*" or "*diversity shift*" in the market.
**DJ2-1**: Data on a market, e.g., position of sale (POS) in a supermarket or stock prices
**DJ2-2**: Data on social events and news
**TJ2-1**: Tangled String or some other tools for explaining a change

Here, a TJ stands for a tool jacket [Hayashi and Ohsawa 2016] where a tool for using data (e.g., a method of AI, data visualization, etc.) is summarized in a form similar to DJ, that is, including the title (e.g., *KeyGraph*), the abstract (e.g., *visualizing the co-occurrence relations between both frequent and infrequent items in the data*), and the input/output variables (*word, item, event, human, time*, etc.). Tangled string (TS) is a method for change explanation, that is, explaining the causality of a noteworthy change, by positioning an event in a string representing a sequence that tangles if an event occurs multiple times [Ohsawa, Hayashi, and Yoshino 2019]. In the sense that there are different time ranges of different trends and these time ranges can be connected by periods bridging a trend and the next trend, the history in the market can be regarded as a string that has some tangled parts within the feature concept *tangled string*.

Change explanation can be logically represented by the predicate "change" defined indirectly in clause (1). This means to position a change as a transition from a trend in a certain period of length $2\Delta t$ if the market changes substantially, as in Eq.(1).

$$| \text{market}(t:t+\Delta t) - \text{market}(t-\Delta t:t)| > Q \qquad (1)$$

$$\text{trend}_1(t:t+\Delta t) :- \text{trend}_2(t-\Delta t:t), \text{change}(t), \qquad (2)$$

$$\text{equal}(\text{market}(t:t+\Delta t), \text{vector}_i) :- \text{trend}_i(t:t+\Delta t) \qquad (3)$$

The cause of the state of the market is given by the changing trend, as in Eq.(2). The requirement is to explain the causality of the change, as in clauses (2) and (3). The value of only the predicate "market," represented by a vector, can be obtained from the data in hand. The latent trend, that is, $\text{trend}_i$ of the $i$-th period in the entire time series, for example $\text{trend}_1$ = hot meal, $\text{trend}_2$ = beverage for cooling the body, are not likely to be included in the data but can be explained by referring to events other than those in the given data. Because such a latent trend may be interpreted using humans' common sense useful for connecting an event in data to other events, one should distinguish change explanation from the *detection* or *prediction* of changes using machine learning technologies (e.g. [Fearnhead and Liu 2007, Hayashi and Yamanishi 2015, Miyaguchi and Yamanishi 2017]). It is essential to create a feature concept for change explanation.

On the other hand, *the diversity shift* proposed by Kahn (1995) as an explanation of essential trend shifts in a market works as a feature concept that matches the requirement Req3 and is computable from data by concretizing as the change in graph-based entropy (GBE [Ohsawa 2018a]). GBE is an index of the diversity of events on their distribution to the subgraphs of a co-occurrence graph, as shown in Eq.(4).

$$Hg = \Sigma_j \, p(\text{cluster}_j) \, \log p(\text{cluster}_j),$$
$$\text{where } p(\text{cluster}_j) = \text{freq}(\text{cluster}_j) / \Sigma_j \, \text{freq}(\text{cluster}_j) \qquad (4)$$

Here, $\text{freq}(\text{cluster}_j)$ denotes the frequency of events in the POS data included in cluster $j$. An *event* here means that some consumers purchased a set of items in a basket, so one basket corresponds to one event. $p(x)$ is the proportion of baskets, including item set $x$. The change in GBE is computed for detecting signs of structural changes in the data and is informative in explaining changes in the trend of consumers' preferences. For the data on the position of sale (POS), this change can be regarded as a sign of the appearance, separation, disappearance, or unification of consumers' interests, regarded as the essence of trend shifts suggested by Kahn. In Fig.6, the bridging edge between the two clusters is cut in the 10[th] week, interpreted as an independent growth of the lower cluster corresponding to spices for cooking stew. The 10[th] week in the data was a hot period in August, but the result of Google Trend supported this, i.e., the frequency of query "stew" (in Japanese) increases every year from mid-August in Japan.

The first author Ohsawa used the DJ store [Hayashi and Ohsawa 2016], a search engine of DJs that lists not only DJs including the query but also DJs used in previous IMDJ sessions for the purpose of satisfying a requirement or a solution corresponding to the query word. He entered the query "explain a precursor of earthquakes," to which "POS data" was shown in the output list of the DJ store. As a result, he noticed the same concept of "diversity shift" used for the POS data in the example above can be reused for the analysis of the distribution of epicenters, and invented the following method to satisfy his requirement.

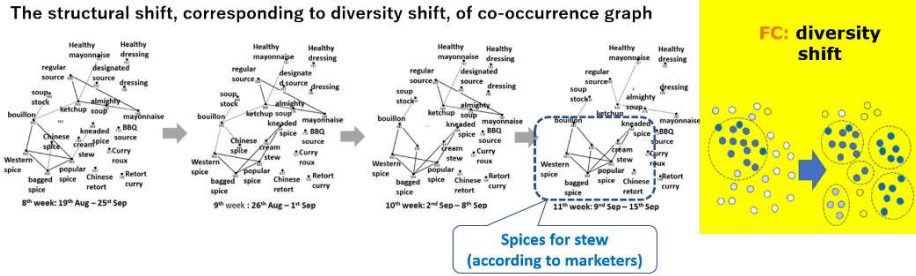

**Fig. 6.** GBE showing the changing point in the market, to aid users' change explanation by the visualization of transitions of graph structures

Fig.7 shows the state transitions in earthquake activation, where in each state from (a) through (c), the appearance of seismic gaps and bridges is regarded as a precursor for a strong earthquake. Thus, the author introduced a simple model to explain the precursory process of earthquakes in two phases:

Phase 1: The diversity of epicenter clusters increases from state (a) or (b)

Phase 2: The clusters above are combined because earthquakes occur in the seismic gap in (c), followed by a large seismic gap in (b).

The entropy defined on the distribution of epicenters into clusters, called the regional entropy on seismic information (RESI [Ohsawa 2018b]), $H$ in Eq.(5), increases in the transition to state (b) from (a) and decreases from (b) to (c). Thus, the saturation in the increase in RESI is supposed to imply a precursory condition of a large earthquake. Here, the diversity shift of epicenters is represented by RESI, borrowing the basic idea of diversity shift in the market represented by GBE. Thus, the common feature concept *diversity shift* or *entropy*, used for marketing, has been diverted by analogy to explaining an earthquake precursor. Here, $S$ is the target region for which RESI is computed. $C_i$ is the $i$-th cluster of the epicenters in $S$.

$$H(S,t) = \sum_i p(C_i|S,t) \log p(C_i|S,t) \qquad (5)$$

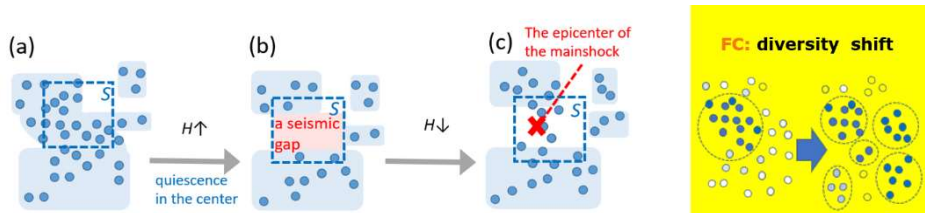

**Fig. 7.** The transition of earthquake activities and its feature concept.

## 4   Feature concepts compared with the pattern language

Feature concepts may be regarded as a customized pattern language, initially proposed by Alexander in urban planning [Alexander et al. 1964, 1977] and diverted so far to systems design (Buschmann et al., 1996). In [Alexander et al., 1977], a set of structural patterns composed of urban elements such as parks, ponds, bridges, and houses were used to explain the existing structures of urban areas and to design new areas. The appearance of these patterns is supposed to be typical in existing urban areas or in contexts where people live and/or work. Each pattern language with an illustration is linked to a context, problem (requirement), and a solution to the problem. The point of keeping such a typical pattern is *at least* twofold: first, the individual thoughts and the communication toward consensus within a team engaged in a task of design or other collaborations can be smoothed by using the patterns as a common language for expressing the contexts, problems, and solutions. Second, the patterns should be connected via relationships from/to each other, which may be hierarchical (class-object) relations or likeliness to be combined. For such a connection, the patterns should be standardized to have a standardized interface with each other.

Similarly, once a feature concept is created and shared with others, it is a typical and useful tool for innovators who think and communicate to federate and/or use data. In addition, the relationships between feature concepts can be, similarly to patterns in a pattern language, the hierarchical structure (e.g., "network of networks" over "network," "diversity shift" over "diversity," etc. where the latter is a part of the former), the connectivity (e.g., connect "diversity shift" and "islands and bridges" for explaining the change in the structure of an organization), etc. More importantly, feature concepts are linked to requirements, as patterns in pattern languages are.

On the other hand, the feature concepts highlight more on the creativity of the users, although the creator of patterns in the pattern language by Alexander or its extensions are also supposed to be creative. The users of feature concepts are encouraged to be radically creative, that is, to create novel feature concepts that may not be accepted by others soon after the initial proposal but still work in individual scientists' thoughts in making a new and goal-oriented algorithm. Furthermore, the elements and structures of a feature concept may be independent of entities in the real world until they are shown to correspond to real entities. Such a correspondence is found when a data scientist embracing the feature concept in his/her mind communicates the individuals having the requirements in real life, or at the time the elements are projected to the attributes (i.e., variables) of data. This point differs from Alexander's pattern language, in which elements correspond to real entities [Alexander 1977], but is partially similar to his abstract formalization of real urbans [Alexander 1964]. This point is similar to the pattern language used in a software design where some activities are defined abstractly, that is, independent of the real behavior of the software or its user(s). Thus, the links from feature concepts to things, events, and actions in the real world should come from the open-minded communication of data scientists with people living in the real world where they declare requirements and ask why and how the requirements are essential.

## 5  Conclusions

Feature concepts have at least two opportunities to contribute to data-federative innovations. The first is in its use as a frame for the pattern or knowledge to be acquired

12from data and as a criterion for evaluating an obtained pattern. The second is that the use of feature concept means to seek novelty in the information to be acquired from data. This means seeking a novel dimension in the performance of data mining, corresponding to Drucker's definition of innovation [Drucker 1985]. For example, the shift from the traditional feature concept *clusters* to *islands and bridges* meant a shift from seeking a believable (statistically significant) pattern to developing a new action scenario that could not be found in the previously used data.

Sharing and using/reusing feature concepts are not only a technology but also a literacy for innovators going beyond available data because the elicited feature concepts may often urge us to collect new data filling the feature concept. In the above example, even if the user finds only *clusters* or *islands* from the original POS, including the purchased items in a supermarket, adding the data on trendy words in a search engine may add the information corresponding to *bridges between islands,* which urges the user to evaluate the result on the criteria for explaining the changes in the market. In this sense, feature concepts are useful in smoothing and improving the quality of the entire process from requirement acquisition, problem solving, data collection, data use/reuse, and the evaluation of the activity using the data. In the near future, we shall compare the processes of creative activities in other domains and use data to divert the ropes of using pattern language by analogy to feature concepts.

**Authors' contributions**

Yukio Ohsawa, Prof.: Conceptualization and exemplification of feature concepts, writing, corresponding author

Sae Kondo, Prof.: Comparison of feature concepts with pattern language

Teruaki Hayashi, Dr.: Development of the DJ store and its applications